\documentclass[10pt,twocolumn,letterpaper]{article}
\usepackage[T1]{fontenc}
\usepackage[utf8]{inputenc}
\usepackage{color}
\usepackage{array}
\usepackage{refstyle}
\usepackage{booktabs}
\usepackage{bm}
\usepackage{multirow}
\usepackage{amsmath}
\usepackage{amssymb}
\usepackage{graphicx}
\usepackage[numbers]{natbib}
\usepackage[unicode=true,
 bookmarks=true,bookmarksnumbered=true,bookmarksopen=false,
 breaklinks=false,pdfborder={0 0 1},backref=false,colorlinks=true]
 {hyperref}

\makeatletter

\AtBeginDocument{\providecommand\figref[1]{\ref{fig:#1}}}
\AtBeginDocument{\providecommand\tblref[1]{\ref{tbl:#1}}}
\AtBeginDocument{\providecommand\secref[1]{\ref{sec:#1}}}
\AtBeginDocument{\providecommand\eqref[1]{\ref{eq:#1}}}
\providecommand{\tabularnewline}{\\}
\newcommand{\lyxdot}{.}

\RS@ifundefined{subsecref}
  {\newref{subsec}{name = \RSsectxt}}
  {}
\RS@ifundefined{thmref}
  {\def\RSthmtxt{theorem~}\newref{thm}{name = \RSthmtxt}}
  {}
\RS@ifundefined{lemref}
  {\def\RSlemtxt{lemma~}\newref{lem}{name = \RSlemtxt}}
  {}

 \usepackage{cvpr}
 \usepackage{times}
 \usepackage{amsmath}
 \usepackage{amssymb}
\cvprfinalcopy

\usepackage{placeins}

\usepackage{color}
\usepackage{xcolor}
\definecolor{darkred}{HTML}{B71C1C}
\definecolor{darkgreen}{HTML}{1B5E20}

\usepackage{pifont}\newcommand{\cmark}{\textcolor{darkgreen}{\ding{51}}}\newcommand{\xmark}{\textcolor{darkred}{\ding{55}}}
\newref{fig}{refcmd={Figure \ref{#1}}}
\newref{tbl}{refcmd={Table \ref{#1}}}
\newref{sec}{refcmd={Section \ref{#1}}}
\newref{eq}{refcmd={Equation \ref{#1}}}

\usepackage{enumitem}

\setlist[description]{leftmargin=0cm}

\@ifundefined{showcaptionsetup}{}{ \PassOptionsToPackage{caption=false}{subfig}}
\usepackage{subfig}
\makeatother

\makeatletter
\begin{document}

\title{Numerical Coordinate Regression with Convolutional Neural Networks}

\author{\begin{tabular}{cccc}
Aiden Nibali & Zhen He & Stuart Morgan & Luke Prendergast\tabularnewline
\multicolumn{4}{c}{La Trobe University, Australia}\tabularnewline
\end{tabular}}
\maketitle
\begin{abstract}
We study deep learning approaches to inferring numerical coordinates
for points of interest in an input image. Existing convolutional neural
network-based solutions to this problem either take a heatmap matching
approach or regress to coordinates with a fully connected output layer.
Neither of these approaches is ideal, since the former is not entirely
differentiable, and the latter lacks inherent spatial generalization.
We propose our differentiable spatial to numerical transform (DSNT)
to fill this gap. The DSNT layer adds no trainable parameters, is
fully differentiable, and exhibits good spatial generalization. Unlike
heatmap matching, DSNT works well with low heatmap resolutions, so
it can be dropped in as an output layer for a wide range of existing
fully convolutional architectures. Consequently, DSNT offers a better
trade-off between inference speed and prediction accuracy compared
to existing techniques. When used to replace the popular heatmap matching
approach used in almost all state-of-the-art methods for pose estimation,
DSNT gives better prediction accuracy for all model architectures
tested.
\end{abstract}
\global\long\def\Var{\operatorname{Var}}
\global\long\def\DSNT{\operatorname{DSNT}}

\section{Introduction}

In recent years, deep convolutional neural networks (CNNs) have proven
to be highly effective general models for a multitude of computer
vision problems \citep{krizhevsky2012imagenet,long2015fully,radford2015unsupervised,newell2016stacked}.
One such problem is \emph{coordinate regression}, where the goal is
to predict a fixed number of location coordinates corresponding to
points of interest in an input image. A well-known instance of this
problem is human pose estimation, for which CNNs are state-of-the-art.
In this paper we study CNN-based solutions to coordinate regression,
using the single-person pose estimation task as an exemplar. Such
solutions may exhibit the desirable properties of spatial generalization
and/or end-to-end differentiability.

Spatial generalization is the ability of a model to generalize knowledge
obtained at one location during training to another at inference time.
If a spatially generalizable model observes a tennis ball in the top-left
of an image during training, it should be able to successfully locate
a similar tennis ball at a previously unseen location in a new image
(\eg the bottom right). It follows that this property will make a
positive contribution to the overall generalization of a coordinate
regression model, since the goal is to find items anywhere in the
image. In general, the success of CNNs is understood to be a result
of the high generalization ability afforded by spatially shared parameters
\citep{lecun1990handwritten}. To maximize this advantage, care must
be taken to avoid trainable layers which can overfit on global structure.
\citet{lin2013nin} note that ``fully connected layers are prone
to overfitting, thus hampering the generalization ability of the overall
network''.

An end-to-end differentiable model can be composed with other differentiable
layers to form a larger model without losing the ability to train
using backpropagation \citep{rumelhart1985backprop}. In the case
of coordinate regression, being end-to-end differentiable means being
able to propagate gradients all the way from the output numerical
coordinates to the input image. It is possible to train a coordinate
regression model without this property, such as by matching predicted
heatmaps to target heatmaps generated from the ground truth locations.
However, this approach cannot be used in architectures where the numerical
coordinates are learned implicitly as intermediate values, including
the prominent example of Spatial Transformer Networks \citep{jaderberg2015spatial}.

There are many CNN-based solutions to other computer vision tasks,
such as classification and semantic segmentation, which exhibit both
spatial generalization and end-to-end differentiability. However,
existing solutions for coordinate regression sacrifice one property
or the other.

\begin{figure}
\begin{centering}
\subfloat[\label{fig:hm_arch}Heatmap matching]{\begin{centering}
\includegraphics{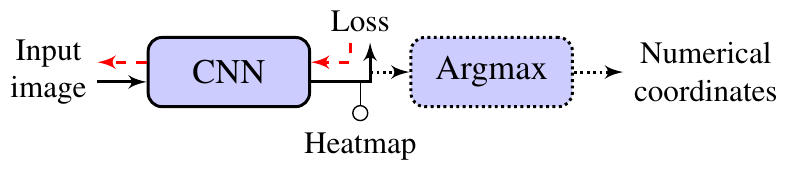}
\par\end{centering}
}
\par\end{centering}
\begin{centering}
\subfloat[Fully connected]{\begin{centering}
\includegraphics{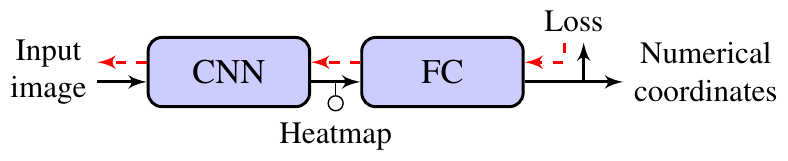}
\par\end{centering}
}
\par\end{centering}
\centering{}\subfloat[DSNT (ours)]{\begin{centering}
\includegraphics{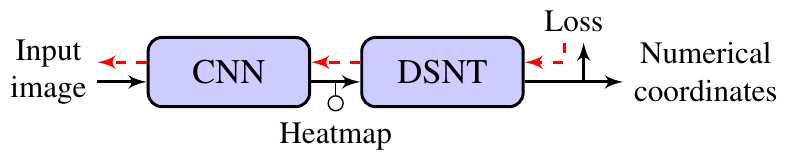}
\par\end{centering}
}\caption{\label{fig:arch_comparison}Comparison of coordinate regression model
architectures. The arrows indicate inference (black) and gradient
flow (dashed red).}
\end{figure}

The most successful existing coordinate regression approach is to
apply a loss directly to output heatmaps rather than numerical coordinates
\citep{tompson2014joint,newell2016stacked,yang2017learning}. Synthetic
heatmaps are generated for each training example by rendering a spherical
2D Gaussian centered on the ground truth coordinates. The model is
trained to produce output images which resemble the synthetic heatmaps
using mean-square-error loss. During inference, numerical coordinates
are obtained from the model's output by computing the argmax of pixel
values, which is a non-differentiable operation. Although this approach
has good spatial generalization, it does have a few disadvantages.
Most notably, gradient flow begins at the heatmap rather than the
numerical coordinates (\figref{hm_arch}). This leads to a disconnect
between the loss function being optimized (similarity between heatmaps)
and the metric we are actually interested in (the distance between
predicted coordinates and ground truth). Only the brightest pixel
is used to calculate numerical coordinates at inference time, but
all of the pixels contribute to the loss during training. Making predictions
based on the argmax also introduces quantization issues, since the
coordinates have their precision tied to the heatmap's resolution.

Another coordinate regression approach is to add a fully connected
layer which produces numerical coordinates \citep{toshev2014deeppose,jaderberg2015spatial}.
An attractive (and sometimes \emph{required}) property of this approach
is that it is possible to backpropagate all the way from the predicted
numerical coordinates to the input image. However, the weights of
the fully-connected layer are highly dependent on the spatial distribution
of the inputs during training. To illustrate this point, consider
an extreme situation where the training set consists entirely of coordinates
located within the left-hand half of the image. Many of the fully
connected layer's input activations will be useless, and as a result
weights corresponding to the right-hand side of the image will not
be trained properly. So although the convolutional part of the model
is spatially invariant, the model as a whole will not generalize well
to objects on the right-hand side of the image. This is an inefficient
usage of the training data, and causes particularly bad performance
on small datasets.

We propose our \emph{differentiable spatial to numerical transform}
(DSNT) layer as an alternative to existing approaches. The DSNT layer
may be used to adapt existing CNN architectures, such as a pretrained
ResNet \citep{he2016resnet}, to coordinate regression problems. Our
technique fully preserves the spatial generalization and end-to-end
differentiability of the model, without introducing additional parameters.
\figref{arch_comparison} illustrates how the DSNT layer fits into
the model as a whole in comparison to fully connected and heatmap
matching approaches. \tblref{desirable_props} summarizes the features
that DSTN poses which selectively appear in fully connected (FC) and
heatmap matching (HM) based approaches.

We find that DSNT is able to consistently outperform the accuracy
of heatmap matching and fully connected approaches across a variety
of architectures on the MPII human pose dataset \citep{andriluka20142d},
and is therefore a suitable replacement in most situations. Our experiments
show that state-of-the-art stacked hourglass models \citep{newell2016stacked}
achieve higher accuracy when heatmap matching is replaced with DSNT.
For ResNet-34 models, DSNT outperforms heatmap matching by 90.5\%
with $7\times7$ pixel heatmaps, and by 2.0\% with $56\times56$ pixel
heatmaps. Since accuracy at low heatmap resolution is much better
with DSNT, a wider variety of efficient architectures may be considered
for coordinate regression. For instance, a simple ResNet-50 network
with DSNT is comparable in accuracy to an 8-stack hourglass network,
but exhibits triple the speed and half of the memory usage during
inference.

\begin{table}
\begin{centering}
\begin{tabular}{|c|>{\centering}p{0.9cm}|>{\centering}p{0.9cm}|>{\centering}p{0.9cm}|}
\hline 
 & HM & FC & DSNT\tabularnewline
\hline 
\hline 
Fully differentiable & \xmark & \cmark & \cmark\tabularnewline
\hline 
Spatially generalizable & \cmark & \xmark & \cmark\tabularnewline
\hline 
No parameters & \cmark & \xmark & \cmark\tabularnewline
\hline 
Good for high-res output & \cmark & \xmark & \cmark\tabularnewline
\hline 
Good for low-res output & \xmark & \cmark & \cmark\tabularnewline
\hline 
Direct coordinate loss & \xmark & \cmark & \cmark\tabularnewline
\hline 
\end{tabular}
\par\end{centering}
\caption{Presence of desirable properties in heatmap matching (HM), fully connected
output (FC), and differentiable spatial to numerical transform (DSNT).}

\label{tbl:desirable_props}
\end{table}

The DSNT layer presented in this paper is very similar to the soft-argmax
operation of \citet{luvizon2017human}, which was developed in parallel
with our own work. The soft-argmax has also been applied to different
problem domains prior to this \citep{yi2016lift,levine2016end}. However,
we extend the idea further by proposing a regularization strategy
which increases prediction accuracy. Additionally, we conduct a comprehensive
set of experiments exploring configurations and properties of the
operation, and the trade-off between accuracy and inference speed
in the context of complete pose estimation models.

\section{Related Work}

Heatmap matching and fully connected layers are prevalent in existing
solutions to problems including human pose estimation and Spatial
Transformer Networks. As such, the following section describes how
existing coordinate regression approaches are applied in those contexts.
Although this paper focuses on pose estimation as an exemplar of the
DSNT layer's capability, our approach is broadly applicable to any
coordinate regression problem.

\subsection{Human pose estimation}

DeepPose \citep{toshev2014deeppose} is one of the earliest CNN-based
models to perform well on the human pose estimation task, and helped
pioneer the current dominance of deep learning in this area. In order
to predict pose joint locations, DeepPose uses a multi-stage cascade
of CNNs with fully connected outputs. The first stage of the cascade
predicts the absolute coordinates of the joint locations, and subsequent
stages refine the predictions by producing relative position deltas.
The authors argue that the cascade arrangement enables reasoning about
human pose at a higher level, since later stages are able to analyze
global structure.

Shortly after DeepPose was published, \citet{tompson2014joint} proposed
a higher accuracy model which uses heatmap matching to calculate loss.
Heatmap matching has since become overwhelmingly dominant amongst
human pose estimation models, including the state-of-the-art stacked
hourglass architecture \citep{newell2016stacked} which is fundamental
to current leaders of the MPII single person pose estimation challenge
\citep{yang2017learning,chen2017advposenet,chou2017self,chu2017multi}.
Each ``hourglass'' in a stacked hourglass network uses the first
half of its layers to downsample activations, and the second half
to upsample back to the original size. By stacking multiple hourglasses
together, the network is able to process data in a repeated bottom-up,
top-down fashion, achieving an effect similar to DeepPose's cascade.
Skip layers are used extensively throughout the architecture, both
within and across individual hourglasses, which makes the model easier
to train with backpropagation.

Very recent research suggests that adversarial training \citep{goodfellow2014generative}
aids in the prediction of likely joint positions by having a discriminator
learn the difference between coherent and nonsensical poses \citep{chen2017advposenet,chou2017self}.
Although we do not conduct such experiments in this paper, we observe
that adversarial training is orthogonal to our findings and could
be combined with our DSNT layer as future work.

\subsection{Spatial Transformer Networks}

The internal Localisation Network component of Spatial Transformer
Networks \citep{jaderberg2015spatial} uses a fully connected layer
to predict translation transformation parameters, which are effectively
just 2D location coordinates. It is not possible to use heatmap matching
in such a model, as gradients must be passed backwards through the
coordinate calculations. In contrast, our DSNT layer could be used
as a drop-in replacement for calculating the translation parameters.

\section{Main idea}

We introduce a new differentiable layer for adapting fully convolutional
networks (FCNs) to coordinate regression. FCNs are a broad class of
CNNs which rely solely on spatially invariant operations to produce
their outputs \citep{lin2013nin}, and are hence naturally spatially
generalizable. Most CNNs with fully connected output layers can be
converted into FCNs by simply removing the fully connected layers.
FCNs are already spatially generalizable and end-to-end differentiable,
so we design our new layer in such a way that these two desirable
properties are preserved. This new layer—which we call the DSNT layer—is
placed at the output of the FCN and transforms spatial heatmaps into
numerical coordinates.

Activations are represented spatially throughout an FCN, which is
very useful for tasks like semantic segmentation \citep{long2015fully}
where the output is intended to be spatial. However, for coordinate
regression tasks like human pose estimation the output needs to be
coordinate pairs. This begs the question: how do we transform spatial
activations into numerical coordinates such that we can still effectively
train the model?

\begin{figure}
\centering{}\noindent\begin{minipage}[t]{1\linewidth}\begin{center}
\begin{minipage}[t]{0.3\linewidth}\begin{center}
\captionsetup[subfigure]{width=1.0\textwidth}\subfloat[Example image and with pose overlay]{\begin{centering}
\hspace{0.09cm}\includegraphics[width=1\linewidth]{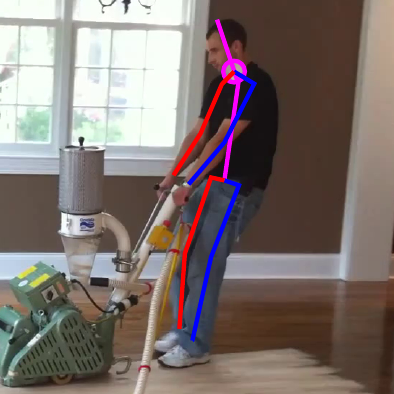}
\par\end{centering}
}
\par\end{center}\end{minipage}\hfill{}\begin{minipage}[t]{0.3\linewidth}\begin{center}
\captionsetup[subfigure]{width=1.0\textwidth}\subfloat[\label{fig:gauss_heatmap}Training target for heatmap matching]{\begin{centering}
\hspace{0.09cm}\includegraphics[width=1\linewidth]{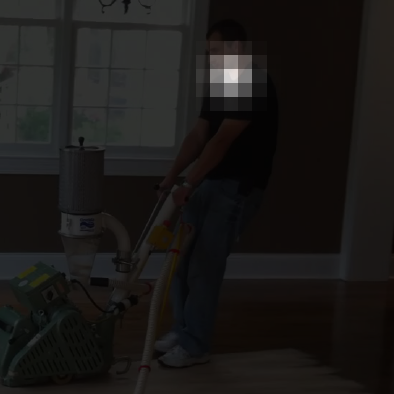}
\par\end{centering}
}
\par\end{center}\end{minipage}\hfill{}\begin{minipage}[t]{0.3\linewidth}\begin{center}
\captionsetup[subfigure]{width=1.0\textwidth}\subfloat[\label{fig:learned_heatmap}Heatmap learned implicitly with DSNT]{\begin{centering}
\hspace{0.09cm}\includegraphics[width=1\linewidth]{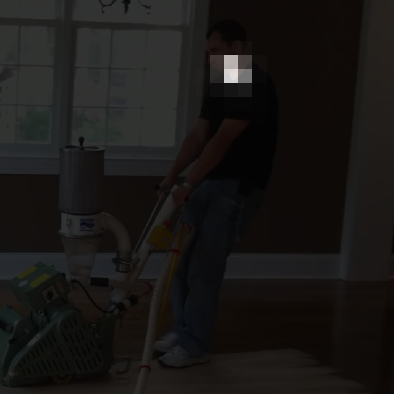}
\par\end{centering}
}
\par\end{center}\end{minipage}\hfill{}
\par\end{center}\end{minipage}\caption{\label{fig:heatmap}Spatial representations of an example neck location.
Image (b) is a 2D Gaussian rendered at the ground truth location,
whereas (c) is learned freely by a model.}
\end{figure}

Consider the case of locating a person's neck in the input image.
This location may be represented spatially as a heatmap (\figref{gauss_heatmap}),
and can be learned by an FCN since it is simply a single-channel image.
The purpose of the DSNT layer is to transform such a heatmap into
numerical coordinates, which is the form of output we require for
coordinate regression. However, we have to be careful about how we
approach designing the DSNT, since we want the layer to be part of
an end-to-end trainable model. For example, if we simply take the
location of the brightest pixel then we cannot calculate meaningful
gradients during training. Therefore, we design the DSNT layer such
that it is able to propagate smooth gradients back through all heatmap
pixels from the numerical coordinates.

In contrast to heatmap matching techniques, we do not require applying
a loss directly to the heatmap output by the FCN to make it resemble
\figref{gauss_heatmap}. Instead, the heatmap is learned indirectly
by optimizing a loss applied to the predicted coordinates output by
the model as a whole. This means that during training the heatmap
will evolve to produce accurate coordinates via the DSNT layer. An
example of an implicitly learned heatmap is shown in \figref{learned_heatmap}.

\section{The Differentiable Spatial to Numerical Transform\label{sec:Calculating-the-DSNT}}

In this section we describe the technical details of our differentiable
spatial to numerical transform (DSNT) layer. The DSNT layer has no
trainable parameters, is fully differentiable, and generalizes spatially.
Accordingly, it is possible to use our layer as part of a CNN model
to enable numerical coordinate outputs without sacrificing end-to-end
learning with backpropagation.

The input to the DSNT is a single-channel normalized heatmap, $\hat{\bm{Z}}$,
represented as an $m\times n$ matrix where $m$ and $n$ correspond
to the heatmap resolution. By ``normalized'' we mean that all elements
of $\hat{\bm{Z}}$ are non-negative and sum to one—the same conditions
which must be fulfilled by a probability distribution. Using such
a normalized heatmap guarantees that predicted coordinates will always
lie within the spatial extent of the heatmap itself. The unnormalized
heatmap output of an FCN, $\bm{Z}$, can be normalized by applying
a heatmap activation function $\hat{\bm{Z}}=\phi(\bm{Z})$. Suitable
choices for $\phi(\bm{Z})$ are discussed in \secref{Heatmap-activation}.

\begin{figure}
\begin{centering}
\includegraphics[width=1\linewidth]{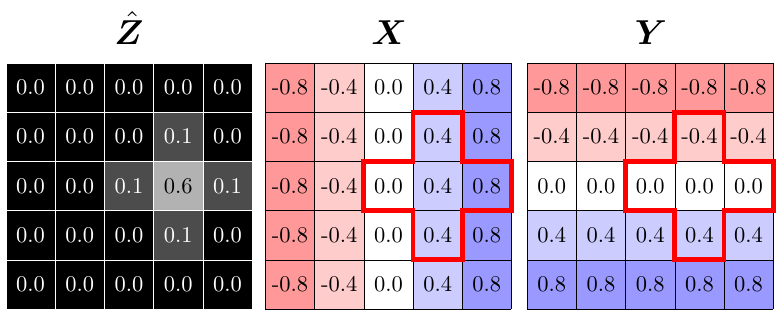}\vspace{-0.5cm}
\par\end{centering}
\begin{centering}
\begin{center}
\resizebox{\linewidth}{!}{$x=\left\langle \hat{\bm{Z}},\bm{X}\right\rangle _{F}=\left(\begin{array}{cccccc}
 &  & 0.1\times0.4 & +\\
0.1\times0.0 & + & 0.6\times0.4 & + & 0.1\times0.8 & +\\
 &  & 0.1\times0.4
\end{array}\right)=0.4$}
\par\end{center}

\begin{center}
\resizebox{\linewidth}{!}{$y=\left\langle \hat{\bm{Z}},\bm{Y}\right\rangle _{F}=\left(\begin{array}{cccccc}
 &  & 0.1\times-0.4 & +\\
0.1\times0.0 & + & 0.6\times0.0 & + & 0.1\times0.0 & +\\
 &  & 0.1\times0.4
\end{array}\right)=0.0$}
\par\end{center}
\par\end{centering}
\centering{}\caption{\label{fig:heatmap_inner_product}Coordinate calculation using the
differentiable spatial to numerical transform (DSNT).}
\end{figure}

Let $\bm{X}$ and $\bm{Y}$ be $m\times n$ matrices, where $X_{i,j}=\frac{2j-(n+1)}{n}$
and $Y_{i,j}=\frac{2i-(m+1)}{m}$. That is, each entry of $\bm{X}$
and $\bm{Y}$ contains its own $x$- or $y$-coordinate respectively,
scaled such that the top-left corner of the image is at $(-1,-1)$
and bottom-right is at $(1,1)$.

By taking a probabilistic interpretation of $\hat{\bm{Z}}$ we can
represent the coordinates, $\bm{\mathrm{c}}$, as a discrete bivariate
random vector with mass function $p(\bm{\mathrm{c}})$ defined as
\[
\operatorname{Pr}(\bm{\mathrm{c}}=\left[\begin{array}{cc}
X_{i,j} & Y_{i,j}\end{array}\right])=\hat{Z}_{i,j}
\]
for all $i=1\ldots m,j=1\ldots n$.

In the heatmap matching approach to coordinate regression, the predicted
numerical coordinates are analogous to the mode of $\bm{\mathrm{c}}$.
For the DSNT layer we instead take our prediction to be the mean of
$\bm{\mathrm{c}}$, denoted $\bm{\mu}=\mathbb{E}[\bm{\mathrm{c}}]$.
Unlike the mode, the mean can a) have its derivative calculated, allowing
us to backpropagate through the DSNT layer; and b) predict coordinates
with sub-pixel precision. \eqref{dsnt} details how the expectation
is calculated, and hence defines the DSNT operation. We use $\left\langle \cdot,\cdot\right\rangle _{F}$
to denote the Frobenius inner product, which is equivalent to taking
the scalar dot product of vectorized matrices.
\begin{equation}
\DSNT(\hat{\bm{Z}})=\bm{\mu}=\left[\begin{array}{cc}
\left\langle \hat{\bm{Z}},\bm{X}\right\rangle {}_{F} & \left\langle \hat{\bm{Z}},\bm{Y}\right\rangle {}_{F}\end{array}\right]\label{eq:dsnt}
\end{equation}

\figref{heatmap_inner_product} illustrates the DSNT operation with
an example. Notice how the symmetrical off-center values of the heatmap
cancel each other out in the calculations. In practice, this property
tends to cause the network to learn heatmaps which are roughly symmetrical
about the predicted location.

One seemingly apparent flaw with using the mean instead of the mode
is that the predicted coordinates will be affected adversely by outliers
in the heatmap. However, it is important to keep in mind that the
heatmap itself is learned with the objective of optimizing coordinate
accuracy. Therefore, during training the model is encouraged to threshold
its activations such that outliers are simply not placed in the heatmap
at all. That is, the network is specifically punished for polluting
the heatmap with low confidence outliers \emph{because} they would
adversely affect results, and hence the model can simply learn to
avoid such situations.

\subsection{Heatmap activation\label{sec:Heatmap-activation}}

As mentioned earlier, a heatmap activation function $\phi(\bm{Z})$
is required to normalize the heatmap before applying the DSNT. Here
we will describe several choices for this function by decomposing
the activation into two parts. Firstly, each element of the input
image $\bm{Z}$ undergoes rectification to produce a non-negative
output. The rectified image $\bm{Z}'$ is then normalized using the
$L^{1}$ norm so that the elements sum to one (\ie $\hat{\bm{Z}}=(\sum Z'_{i,j})^{-1}\bm{Z}'$).

\begin{table}
\centering{}\caption{\label{tbl:heatmap_rectification}Heatmap activation functions and
their corresponding human pose estimation results.}
 \begin{tabular}{lll}
\toprule 
Name & Rectification & PCKh\tabularnewline
\midrule 
\textbf{Softmax} & $Z'_{i,j}=\exp(Z_{i,j})$ & \textbf{86.81\%}\tabularnewline
Abs & $Z'_{i,j}=\left|Z_{i,j}\right|$ & 86.48\%\tabularnewline
ReLU & $Z'_{i,j}=\max(0,Z_{i,j})$ & 86.69\%\tabularnewline
Sigmoid & $Z'_{i,j}=(1+\exp(-Z_{i,j}))^{-1}$ & 86.71\%\tabularnewline
\bottomrule
\end{tabular}
\end{table}

\tblref{heatmap_rectification} shows some possible options for the
rectification function, along with validation set PCKh accuracy measurements
on the MPII human pose dataset. These results were gathered using
ResNet-34 models pretrained on ImageNet, dilated to produce a heatmap
resolution of $28\times28$ pixels. No regularization was used. Although
the choice of rectification function does not appear to have a large
impact on results, our experiments indicate that softmax works best.

\section{Loss function}

Since the DSNT layer outputs numerical coordinates, it is possible
to directly calculate the two-dimensional Euclidean distance between
the prediction $\bm{\mu}$ and ground truth $\bm{p}$. We take advantage
of this fact to formulate the core term of our loss function (\eqref{euclidean_loss}).
\begin{equation}
\mathcal{L}_{euc}(\bm{\mu},\bm{p})=\left\Vert \bm{p}-\bm{\mu}\right\Vert _{2}\label{eq:euclidean_loss}
\end{equation}

The Euclidean loss function has the advantage of directly optimizing
the metric we are interested in: the distance between the predicted
and actual locations.

\begin{figure}
\centering{}\includegraphics[width=1\linewidth]{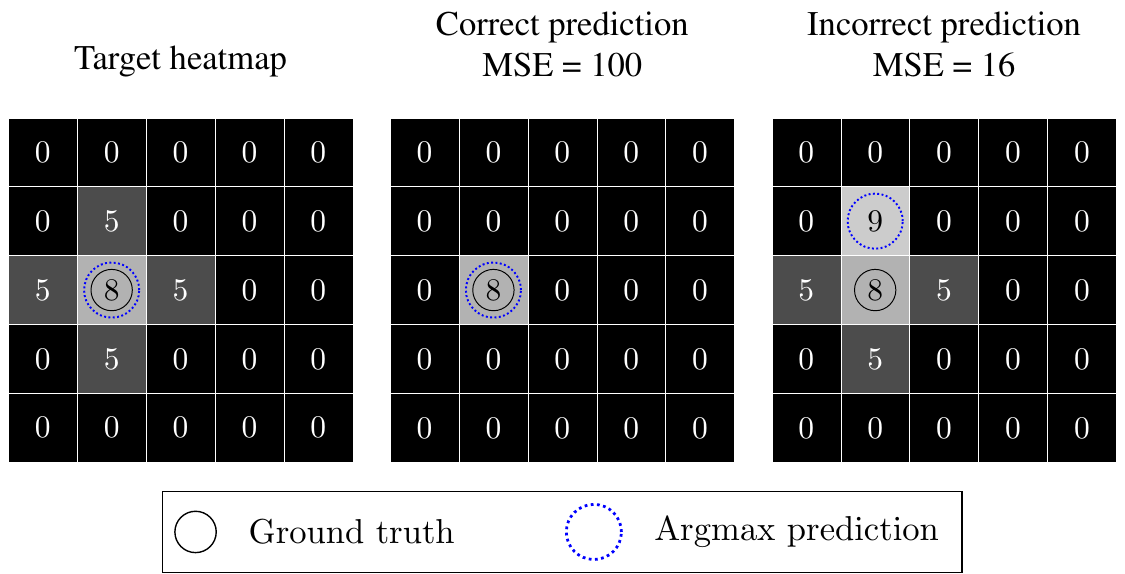}\caption{\label{fig:mse-problem}When heatmap matching, it is possible for
predictions to worsen despite the pixel-wise MSE improving.}
\end{figure}

Contrast this with the mean-square-error (MSE) loss used in heatmap
matching, which optimizes the pixel-wise similarity between the output
and a synthetic heatmap generated from ground truth locations. The
pixel-wise MSE loss is a much less direct way of optimizing the metric
that we actually care about. During training, the model is completely
ignorant of the fact that coordinate predictions are based solely
on the brightest heatmap pixel. Another way to put this is that despite
the Euclidean loss having a global minimum when the MSE loss is zero,
we aren't guaranteed that an optimization step which improves MSE
loss will improve our results. \figref{mse-problem} illustrates an
example situation where improving the MSE loss degrades the predictive
accuracy of the model. In this case we see that the output with a
single pixel at the correct location has worse MSE but better location
prediction than an almost perfectly matching heatmap with the brightest
pixel placed incorrectly.

\subsection{Regularization}

There are many different possible heatmaps that will lead to the same
coordinates being output from the DSNT layer. For example, the spread
of the heatmap has no effect on the output—blobs resembling 2D Gaussians
with large variance and small variance can produce identical coordinates.
Although such freedom may be viewed as beneficial, a potential drawback
is that the model does not have strongly supervised pixel-wise gradients
through the heatmap during training. Experimentally, we find that
providing such supervision via regularization can yield marked performance
improvements over vanilla DSNT.

\eqref{combined_loss} shows how regularization is incorporated into
the DSNT loss function. A regularization coefficient, $\lambda$,
is used to set the strength of the regularizer, $\mathcal{\mathcal{L}}_{reg}$.
\begin{equation}
\mathcal{L}(\hat{\bm{Z}},\bm{p})=\mathcal{L}_{euc}(\DSNT(\hat{\bm{Z}}),\bm{p})+\lambda\mathcal{L}_{reg}(\hat{\bm{Z}})\label{eq:combined_loss}
\end{equation}

\subsubsection{Variance regularization}

By expanding upon the probabilistic interpretation of the DSNT layer
(\secref{Calculating-the-DSNT}), we can calculate the variance of
coordinates. This is described for $x$-coordinates in \eqref{variance}
($y$-coordinates are handled similarly). The calculated variance
represents the ``spread'' of the blob in the heatmap, which is analogous
to the size of the synthetic 2D Gaussian drawn in the heatmap matching
approach.
\begin{eqnarray}
\Var[\mathrm{c}_{x}] & = & \mathbb{E}[(\mathrm{c}_{x}-\mathbb{E}[\mathrm{c}_{x}])^{2}]\label{eq:variance}\\
 & = & \left\langle \hat{\bm{Z}},(\bm{X}-\mu_{x})\odot(\bm{X}-\mu_{x})\right\rangle _{F}\nonumber 
\end{eqnarray}

We are now able to introduce a variance regularization term, \eqref{reg-var}.
The ``spread'' of the learned heatmaps is controlled by a hyperparameter,
the target variance, $\sigma_{t}^{2}$. Note that this regularization
term does not directly constrain the specific shape of learned heatmaps.
\begin{equation}
\mathcal{L}_{var}(\hat{\bm{Z}})=(\Var[\mathrm{c}_{x}]-\sigma_{t}^{2})^{2}+(\Var[\mathrm{c}_{y}]-\sigma_{t}^{2})^{2}\label{eq:reg-var}
\end{equation}

\subsubsection{Distribution regularization}

Alternatively, we can impose even stricter regularization on the appearance
of the heatmap to directly encourage a certain shape. More specifically,
to force the heatmap to resemble a spherical Gaussian, we can minimize
the divergence between the generated heatmap and an appropriate target
normal distribution. \eqref{reg-divergence} defines the distribution
regularization term, where $D(\cdot||\cdot)$ is a divergence measure
(\eg Jensen-Shannon divergence).
\begin{equation}
\mathcal{L}_{D}(\hat{\bm{Z}},\bm{p})=D(p(\bm{\mathrm{c}})||\mathcal{N}(\bm{p},\sigma_{t}^{2}\bm{I}_{2}))\label{eq:reg-divergence}
\end{equation}

Adding a regularization term of this form is similar to incorporating
the usual heatmap matching objective into the DSNT loss function.

\subsubsection*{Selecting the best regularization}

\begin{table}
\centering{}\caption{\label{tbl:reg_results}Pose estimation results for different regularization
terms, using a ResNet-34@28px model.}
 \begin{tabular}{llcc}
\toprule 
\multirow{2}{*}{Regularization} & \multirow{2}{*}{$\lambda$} & \multicolumn{2}{c}{Validation PCKh}\tabularnewline
 &  & $\sigma_{t}=1$ & $\sigma_{t}=2$\tabularnewline
\midrule
None & N/A & \multicolumn{2}{c}{86.86\%}\tabularnewline
Variance & 100 & 84.58\% & 85.88\%\tabularnewline
Kullback-Leibler & 1 & 84.67\% & 84.15\%\tabularnewline
\textbf{Jensen-Shannon} & \textbf{1} & \textbf{87.59\%} & \textbf{86.71\%}\tabularnewline
\bottomrule
\end{tabular}
\end{table}

\begin{figure}
\begin{centering}
\includegraphics[width=1\linewidth]{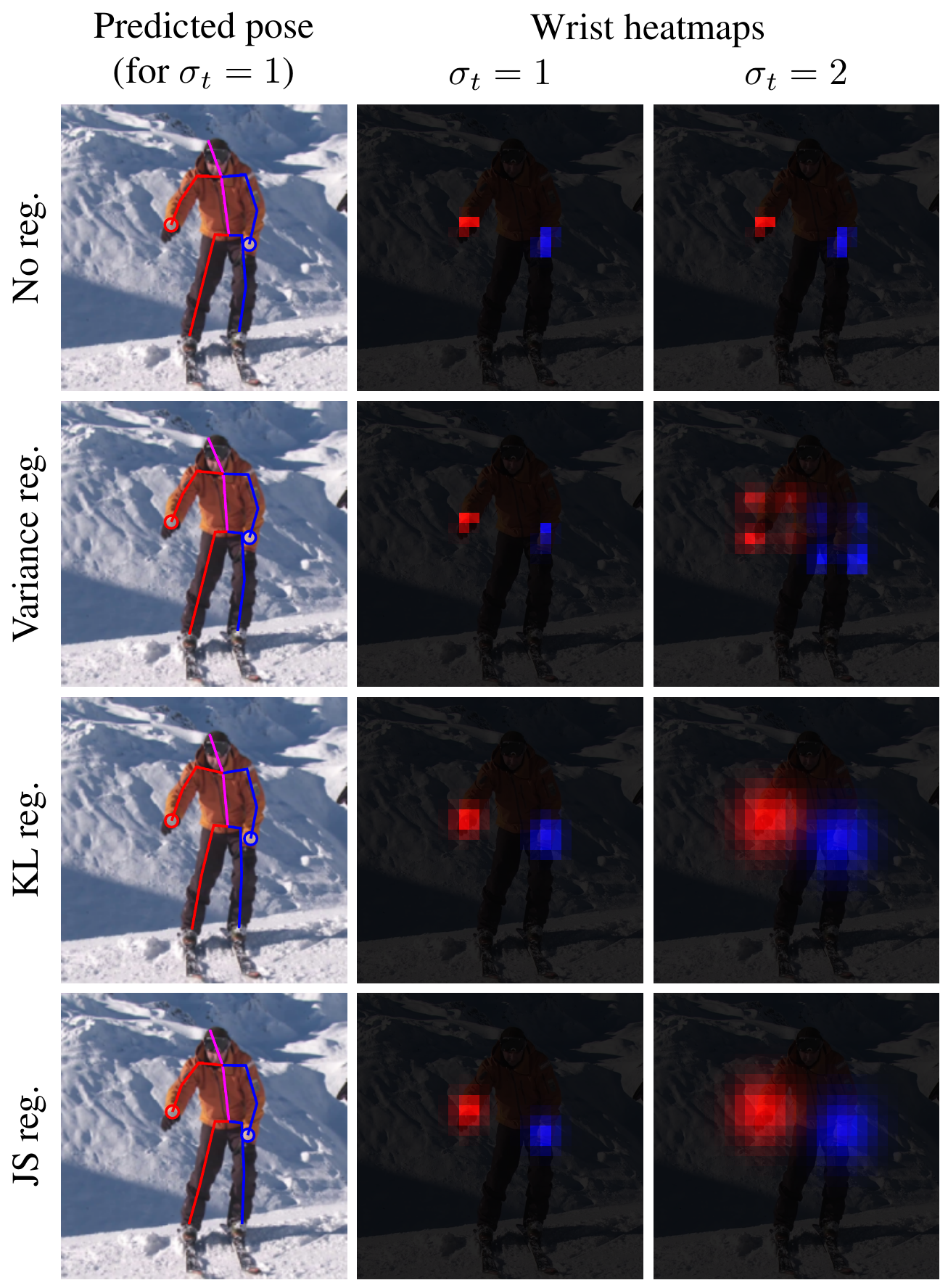}
\par\end{centering}
\caption{\label{fig:reg_hm_appearance}Heatmap appearance for models trained
with different regularization terms (red = right wrist, blue = left
wrist).}
\end{figure}

\begin{figure}
\centering{}\clearpage{}\begingroup
  \makeatletter
  \providecommand\color[2][]{    \GenericError{(gnuplot) \space\space\space\@spaces}{      Package color not loaded in conjunction with
      terminal option `colourtext'    }{See the gnuplot documentation for explanation.    }{Either use 'blacktext' in gnuplot or load the package
      color.sty in LaTeX.}    \renewcommand\color[2][]{}  }  \providecommand\includegraphics[2][]{    \GenericError{(gnuplot) \space\space\space\@spaces}{      Package graphicx or graphics not loaded    }{See the gnuplot documentation for explanation.    }{The gnuplot epslatex terminal needs graphicx.sty or graphics.sty.}    \renewcommand\includegraphics[2][]{}  }  \providecommand\rotatebox[2]{#2}  \@ifundefined{ifGPcolor}{    \newif\ifGPcolor
    \GPcolortrue
  }{}  \@ifundefined{ifGPblacktext}{    \newif\ifGPblacktext
    \GPblacktexttrue
  }{}    \let\gplgaddtomacro\g@addto@macro
    \gdef\gplbacktext{}  \gdef\gplfronttext{}  \makeatother
  \ifGPblacktext
        \def\colorrgb#1{}    \def\colorgray#1{}  \else
        \ifGPcolor
      \def\colorrgb#1{\color[rgb]{#1}}      \def\colorgray#1{\color[gray]{#1}}      \expandafter\def\csname LTw\endcsname{\color{white}}      \expandafter\def\csname LTb\endcsname{\color{black}}      \expandafter\def\csname LTa\endcsname{\color{black}}      \expandafter\def\csname LT0\endcsname{\color[rgb]{1,0,0}}      \expandafter\def\csname LT1\endcsname{\color[rgb]{0,1,0}}      \expandafter\def\csname LT2\endcsname{\color[rgb]{0,0,1}}      \expandafter\def\csname LT3\endcsname{\color[rgb]{1,0,1}}      \expandafter\def\csname LT4\endcsname{\color[rgb]{0,1,1}}      \expandafter\def\csname LT5\endcsname{\color[rgb]{1,1,0}}      \expandafter\def\csname LT6\endcsname{\color[rgb]{0,0,0}}      \expandafter\def\csname LT7\endcsname{\color[rgb]{1,0.3,0}}      \expandafter\def\csname LT8\endcsname{\color[rgb]{0.5,0.5,0.5}}    \else
            \def\colorrgb#1{\color{black}}      \def\colorgray#1{\color[gray]{#1}}      \expandafter\def\csname LTw\endcsname{\color{white}}      \expandafter\def\csname LTb\endcsname{\color{black}}      \expandafter\def\csname LTa\endcsname{\color{black}}      \expandafter\def\csname LT0\endcsname{\color{black}}      \expandafter\def\csname LT1\endcsname{\color{black}}      \expandafter\def\csname LT2\endcsname{\color{black}}      \expandafter\def\csname LT3\endcsname{\color{black}}      \expandafter\def\csname LT4\endcsname{\color{black}}      \expandafter\def\csname LT5\endcsname{\color{black}}      \expandafter\def\csname LT6\endcsname{\color{black}}      \expandafter\def\csname LT7\endcsname{\color{black}}      \expandafter\def\csname LT8\endcsname{\color{black}}    \fi
  \fi
    \setlength{\unitlength}{0.0500bp}    \ifx\gptboxheight\undefined      \newlength{\gptboxheight}      \newlength{\gptboxwidth}      \newsavebox{\gptboxtext}    \fi    \setlength{\fboxrule}{0.5pt}    \setlength{\fboxsep}{1pt}\begin{picture}(4700.00,2260.00)    \gplgaddtomacro\gplbacktext{      \csname LTb\endcsname      \put(727,595){\makebox(0,0)[r]{\strut{}85\%}}      \csname LTb\endcsname      \put(727,1082){\makebox(0,0)[r]{\strut{}86\%}}      \csname LTb\endcsname      \put(727,1568){\makebox(0,0)[r]{\strut{}87\%}}      \csname LTb\endcsname      \put(727,2055){\makebox(0,0)[r]{\strut{}88\%}}      \csname LTb\endcsname      \put(937,409){\makebox(0,0){\strut{}$0.5$}}      \csname LTb\endcsname      \put(1479,409){\makebox(0,0){\strut{}$1$}}      \csname LTb\endcsname      \put(2021,409){\makebox(0,0){\strut{}$2$}}      \csname LTb\endcsname      \put(2563,409){\makebox(0,0){\strut{}$4$}}    }    \gplgaddtomacro\gplfronttext{      \csname LTb\endcsname      \put(124,1325){\rotatebox{-270}{\makebox(0,0){\strut{}PCKh total}}}      \csname LTb\endcsname      \put(1750,130){\makebox(0,0){\strut{}$\sigma_t$ (px)}}      \csname LTb\endcsname      \put(1662,762){\makebox(0,0)[r]{\strut{}$\lambda=1.0$}}    }    \gplgaddtomacro\gplbacktext{      \csname LTb\endcsname      \put(2662,595){\makebox(0,0)[r]{\strut{}}}      \csname LTb\endcsname      \put(2662,1082){\makebox(0,0)[r]{\strut{}}}      \csname LTb\endcsname      \put(2662,1568){\makebox(0,0)[r]{\strut{}}}      \csname LTb\endcsname      \put(2662,2055){\makebox(0,0)[r]{\strut{}}}      \csname LTb\endcsname      \put(2885,409){\makebox(0,0){\strut{}$0.2$}}      \csname LTb\endcsname      \put(3561,409){\makebox(0,0){\strut{}$1$}}      \csname LTb\endcsname      \put(4238,409){\makebox(0,0){\strut{}$5$}}      \csname LTb\endcsname      \put(4529,409){\makebox(0,0){\strut{}$10$}}    }    \gplgaddtomacro\gplfronttext{      \csname LTb\endcsname      \put(3685,130){\makebox(0,0){\strut{}$\lambda$}}      \csname LTb\endcsname      \put(3648,762){\makebox(0,0)[r]{\strut{}$\sigma_t=1.0$}}    }    \gplbacktext
    \put(0,0){\includegraphics{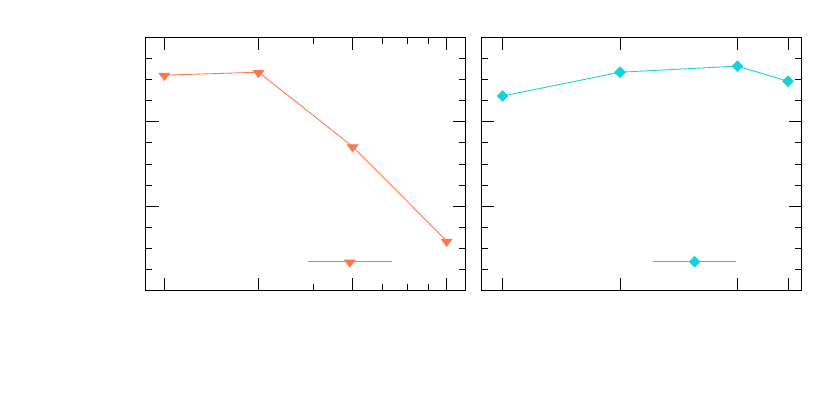}}    \gplfronttext
  \end{picture}\endgroup
\clearpage{}\vspace{-0.5cm}
\caption{\label{fig:reg_params}Varying the Gaussian size and regularization
strength for JS regularization.}
\end{figure}

In order to determine the best performing regularization term, we
conducted a series of experiments on the MPII human pose dataset with
a ResNet-34@28px model.

Firstly, we compared different options for the regularization function,
$\mathcal{L}_{reg}$: variance regularization, and distribution regularization
with Kullback-Leibler (KL) and Jensen-Shannon (JS) divergences. The
pose estimation results in \tblref{reg_results} indicate that JS
distribution regularization achieves the highest accuracy. The sample
heatmap images shown in \figref{reg_hm_appearance} illustrate how
dramatically the choice of regularization term can change the appearance
of heatmaps. For example, distribution regularization (using either
KL or JS divergence) very effectively encourages the production of
distinctly Gaussian-shaped blobs. In contrast, variance regularization
with $\sigma_{t}=2$ results in an interesting strategy of splitting
the heatmap into four blobs around the joint.

We conducted further experiments to determine the optimal regularization
hyperparameters (\figref{reg_params}). The accuracy of the model
was found to be quite robust with respect to the regularization strength,
$\lambda$ (\eqref{combined_loss}). In terms of the target Gaussian
standard deviation, $\sigma_{t}$, values in the range of half a pixel
to one pixel were found to work well.

\section{Experiments\label{sec:Experimental-Results}}

\begin{figure*}
\begin{minipage}[t]{0.32\linewidth}\begin{center}
\clearpage{}\begingroup
  \makeatletter
  \providecommand\color[2][]{    \GenericError{(gnuplot) \space\space\space\@spaces}{      Package color not loaded in conjunction with
      terminal option `colourtext'    }{See the gnuplot documentation for explanation.    }{Either use 'blacktext' in gnuplot or load the package
      color.sty in LaTeX.}    \renewcommand\color[2][]{}  }  \providecommand\includegraphics[2][]{    \GenericError{(gnuplot) \space\space\space\@spaces}{      Package graphicx or graphics not loaded    }{See the gnuplot documentation for explanation.    }{The gnuplot epslatex terminal needs graphicx.sty or graphics.sty.}    \renewcommand\includegraphics[2][]{}  }  \providecommand\rotatebox[2]{#2}  \@ifundefined{ifGPcolor}{    \newif\ifGPcolor
    \GPcolortrue
  }{}  \@ifundefined{ifGPblacktext}{    \newif\ifGPblacktext
    \GPblacktexttrue
  }{}    \let\gplgaddtomacro\g@addto@macro
    \gdef\gplbacktext{}  \gdef\gplfronttext{}  \makeatother
  \ifGPblacktext
        \def\colorrgb#1{}    \def\colorgray#1{}  \else
        \ifGPcolor
      \def\colorrgb#1{\color[rgb]{#1}}      \def\colorgray#1{\color[gray]{#1}}      \expandafter\def\csname LTw\endcsname{\color{white}}      \expandafter\def\csname LTb\endcsname{\color{black}}      \expandafter\def\csname LTa\endcsname{\color{black}}      \expandafter\def\csname LT0\endcsname{\color[rgb]{1,0,0}}      \expandafter\def\csname LT1\endcsname{\color[rgb]{0,1,0}}      \expandafter\def\csname LT2\endcsname{\color[rgb]{0,0,1}}      \expandafter\def\csname LT3\endcsname{\color[rgb]{1,0,1}}      \expandafter\def\csname LT4\endcsname{\color[rgb]{0,1,1}}      \expandafter\def\csname LT5\endcsname{\color[rgb]{1,1,0}}      \expandafter\def\csname LT6\endcsname{\color[rgb]{0,0,0}}      \expandafter\def\csname LT7\endcsname{\color[rgb]{1,0.3,0}}      \expandafter\def\csname LT8\endcsname{\color[rgb]{0.5,0.5,0.5}}    \else
            \def\colorrgb#1{\color{black}}      \def\colorgray#1{\color[gray]{#1}}      \expandafter\def\csname LTw\endcsname{\color{white}}      \expandafter\def\csname LTb\endcsname{\color{black}}      \expandafter\def\csname LTa\endcsname{\color{black}}      \expandafter\def\csname LT0\endcsname{\color{black}}      \expandafter\def\csname LT1\endcsname{\color{black}}      \expandafter\def\csname LT2\endcsname{\color{black}}      \expandafter\def\csname LT3\endcsname{\color{black}}      \expandafter\def\csname LT4\endcsname{\color{black}}      \expandafter\def\csname LT5\endcsname{\color{black}}      \expandafter\def\csname LT6\endcsname{\color{black}}      \expandafter\def\csname LT7\endcsname{\color{black}}      \expandafter\def\csname LT8\endcsname{\color{black}}    \fi
  \fi
    \setlength{\unitlength}{0.0500bp}    \ifx\gptboxheight\undefined      \newlength{\gptboxheight}      \newlength{\gptboxwidth}      \newsavebox{\gptboxtext}    \fi    \setlength{\fboxrule}{0.5pt}    \setlength{\fboxsep}{1pt}\begin{picture}(3440.00,2540.00)    \gplgaddtomacro\gplbacktext{      \csname LTb\endcsname      \put(747,769){\makebox(0,0)[r]{\strut{}80\%}}      \csname LTb\endcsname      \put(747,1117){\makebox(0,0)[r]{\strut{}82\%}}      \csname LTb\endcsname      \put(747,1465){\makebox(0,0)[r]{\strut{}84\%}}      \csname LTb\endcsname      \put(747,1813){\makebox(0,0)[r]{\strut{}86\%}}      \csname LTb\endcsname      \put(747,2161){\makebox(0,0)[r]{\strut{}88\%}}      \csname LTb\endcsname      \put(920,409){\makebox(0,0){\strut{}$7$}}      \csname LTb\endcsname      \put(1634,409){\makebox(0,0){\strut{}$14$}}      \csname LTb\endcsname      \put(2348,409){\makebox(0,0){\strut{}$28$}}      \csname LTb\endcsname      \put(3062,409){\makebox(0,0){\strut{}$56$}}    }    \gplgaddtomacro\gplfronttext{      \csname LTb\endcsname      \put(144,1465){\rotatebox{-270}{\makebox(0,0){\strut{}PCKh total}}}      \csname LTb\endcsname      \put(1991,130){\makebox(0,0){\strut{}Heatmap resolution (px)}}      \csname LTb\endcsname      \put(2345,1320){\makebox(0,0)[r]{\strut{}\footnotesize HM}}      \csname LTb\endcsname      \put(2345,1134){\makebox(0,0)[r]{\strut{}\footnotesize FC}}      \csname LTb\endcsname      \put(2345,948){\makebox(0,0)[r]{\strut{}\footnotesize DSNT}}      \csname LTb\endcsname      \put(2345,762){\makebox(0,0)[r]{\strut{}\footnotesize DSNTr}}    }    \gplbacktext
    \put(0,0){\includegraphics{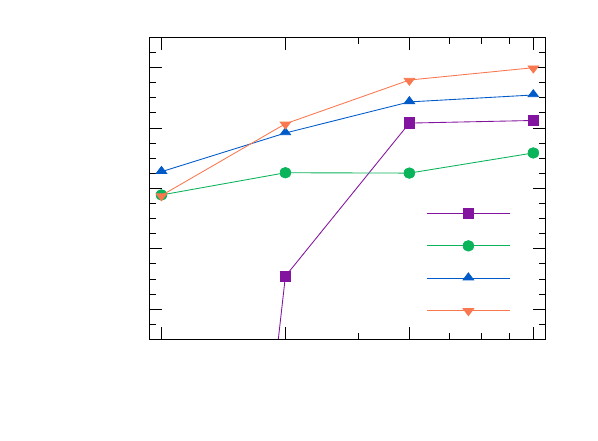}}    \gplfronttext
  \end{picture}\endgroup
\clearpage{}\vspace{-0.8cm}
\par\end{center}
\caption{\label{fig:outstrat}Varying output resolution and strategy for ResNet-34
models.}
\end{minipage}\hfill{}\begin{minipage}[t]{0.32\linewidth}\begin{center}
\clearpage{}\begingroup
  \makeatletter
  \providecommand\color[2][]{    \GenericError{(gnuplot) \space\space\space\@spaces}{      Package color not loaded in conjunction with
      terminal option `colourtext'    }{See the gnuplot documentation for explanation.    }{Either use 'blacktext' in gnuplot or load the package
      color.sty in LaTeX.}    \renewcommand\color[2][]{}  }  \providecommand\includegraphics[2][]{    \GenericError{(gnuplot) \space\space\space\@spaces}{      Package graphicx or graphics not loaded    }{See the gnuplot documentation for explanation.    }{The gnuplot epslatex terminal needs graphicx.sty or graphics.sty.}    \renewcommand\includegraphics[2][]{}  }  \providecommand\rotatebox[2]{#2}  \@ifundefined{ifGPcolor}{    \newif\ifGPcolor
    \GPcolortrue
  }{}  \@ifundefined{ifGPblacktext}{    \newif\ifGPblacktext
    \GPblacktexttrue
  }{}    \let\gplgaddtomacro\g@addto@macro
    \gdef\gplbacktext{}  \gdef\gplfronttext{}  \makeatother
  \ifGPblacktext
        \def\colorrgb#1{}    \def\colorgray#1{}  \else
        \ifGPcolor
      \def\colorrgb#1{\color[rgb]{#1}}      \def\colorgray#1{\color[gray]{#1}}      \expandafter\def\csname LTw\endcsname{\color{white}}      \expandafter\def\csname LTb\endcsname{\color{black}}      \expandafter\def\csname LTa\endcsname{\color{black}}      \expandafter\def\csname LT0\endcsname{\color[rgb]{1,0,0}}      \expandafter\def\csname LT1\endcsname{\color[rgb]{0,1,0}}      \expandafter\def\csname LT2\endcsname{\color[rgb]{0,0,1}}      \expandafter\def\csname LT3\endcsname{\color[rgb]{1,0,1}}      \expandafter\def\csname LT4\endcsname{\color[rgb]{0,1,1}}      \expandafter\def\csname LT5\endcsname{\color[rgb]{1,1,0}}      \expandafter\def\csname LT6\endcsname{\color[rgb]{0,0,0}}      \expandafter\def\csname LT7\endcsname{\color[rgb]{1,0.3,0}}      \expandafter\def\csname LT8\endcsname{\color[rgb]{0.5,0.5,0.5}}    \else
            \def\colorrgb#1{\color{black}}      \def\colorgray#1{\color[gray]{#1}}      \expandafter\def\csname LTw\endcsname{\color{white}}      \expandafter\def\csname LTb\endcsname{\color{black}}      \expandafter\def\csname LTa\endcsname{\color{black}}      \expandafter\def\csname LT0\endcsname{\color{black}}      \expandafter\def\csname LT1\endcsname{\color{black}}      \expandafter\def\csname LT2\endcsname{\color{black}}      \expandafter\def\csname LT3\endcsname{\color{black}}      \expandafter\def\csname LT4\endcsname{\color{black}}      \expandafter\def\csname LT5\endcsname{\color{black}}      \expandafter\def\csname LT6\endcsname{\color{black}}      \expandafter\def\csname LT7\endcsname{\color{black}}      \expandafter\def\csname LT8\endcsname{\color{black}}    \fi
  \fi
    \setlength{\unitlength}{0.0500bp}    \ifx\gptboxheight\undefined      \newlength{\gptboxheight}      \newlength{\gptboxwidth}      \newsavebox{\gptboxtext}    \fi    \setlength{\fboxrule}{0.5pt}    \setlength{\fboxsep}{1pt}\begin{picture}(3440.00,2540.00)    \gplgaddtomacro\gplbacktext{      \csname LTb\endcsname      \put(747,753){\makebox(0,0)[r]{\strut{}84\%}}      \csname LTb\endcsname      \put(747,1070){\makebox(0,0)[r]{\strut{}85\%}}      \csname LTb\endcsname      \put(747,1386){\makebox(0,0)[r]{\strut{}86\%}}      \csname LTb\endcsname      \put(747,1702){\makebox(0,0)[r]{\strut{}87\%}}      \csname LTb\endcsname      \put(747,2019){\makebox(0,0)[r]{\strut{}88\%}}      \csname LTb\endcsname      \put(747,2335){\makebox(0,0)[r]{\strut{}89\%}}      \csname LTb\endcsname      \put(1032,409){\makebox(0,0){\strut{}$18$}}      \csname LTb\endcsname      \put(1397,409){\makebox(0,0){\strut{}$34$}}      \csname LTb\endcsname      \put(1763,409){\makebox(0,0){\strut{}$50$}}      \csname LTb\endcsname      \put(2927,409){\makebox(0,0){\strut{}$101$}}    }    \gplgaddtomacro\gplfronttext{      \csname LTb\endcsname      \put(144,1465){\rotatebox{-270}{\makebox(0,0){\strut{}PCKh total}}}      \csname LTb\endcsname      \put(1991,130){\makebox(0,0){\strut{}Depth (ResNet-$x$)}}      \csname LTb\endcsname      \put(2331,1320){\makebox(0,0){\strut{}Resolution}}      \csname LTb\endcsname      \put(2345,1134){\makebox(0,0)[r]{\strut{}\footnotesize 14 px}}      \csname LTb\endcsname      \put(2345,948){\makebox(0,0)[r]{\strut{}\footnotesize 28 px}}      \csname LTb\endcsname      \put(2345,762){\makebox(0,0)[r]{\strut{}\footnotesize 56 px}}    }    \gplbacktext
    \put(0,0){\includegraphics{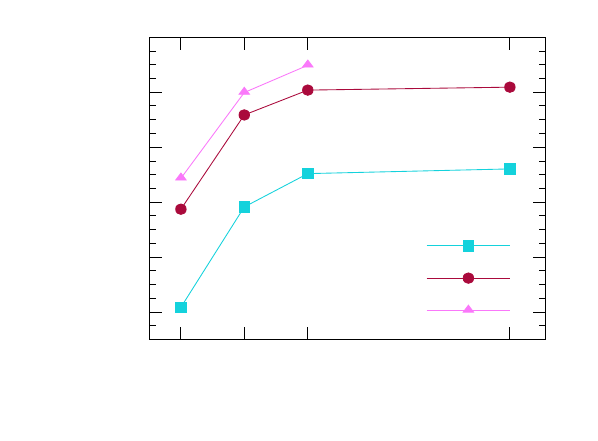}}    \gplfronttext
  \end{picture}\endgroup
\clearpage{}\vspace{-0.8cm}
\par\end{center}
\caption{\label{fig:depth}Varying ResNet \citep{he2016resnet} depth and heatmap
resolution for DSNTr.}
\end{minipage}\hfill{}\begin{minipage}[t]{0.32\linewidth}\begin{center}
\clearpage{}\begingroup
  \makeatletter
  \providecommand\color[2][]{    \GenericError{(gnuplot) \space\space\space\@spaces}{      Package color not loaded in conjunction with
      terminal option `colourtext'    }{See the gnuplot documentation for explanation.    }{Either use 'blacktext' in gnuplot or load the package
      color.sty in LaTeX.}    \renewcommand\color[2][]{}  }  \providecommand\includegraphics[2][]{    \GenericError{(gnuplot) \space\space\space\@spaces}{      Package graphicx or graphics not loaded    }{See the gnuplot documentation for explanation.    }{The gnuplot epslatex terminal needs graphicx.sty or graphics.sty.}    \renewcommand\includegraphics[2][]{}  }  \providecommand\rotatebox[2]{#2}  \@ifundefined{ifGPcolor}{    \newif\ifGPcolor
    \GPcolortrue
  }{}  \@ifundefined{ifGPblacktext}{    \newif\ifGPblacktext
    \GPblacktexttrue
  }{}    \let\gplgaddtomacro\g@addto@macro
    \gdef\gplbacktext{}  \gdef\gplfronttext{}  \makeatother
  \ifGPblacktext
        \def\colorrgb#1{}    \def\colorgray#1{}  \else
        \ifGPcolor
      \def\colorrgb#1{\color[rgb]{#1}}      \def\colorgray#1{\color[gray]{#1}}      \expandafter\def\csname LTw\endcsname{\color{white}}      \expandafter\def\csname LTb\endcsname{\color{black}}      \expandafter\def\csname LTa\endcsname{\color{black}}      \expandafter\def\csname LT0\endcsname{\color[rgb]{1,0,0}}      \expandafter\def\csname LT1\endcsname{\color[rgb]{0,1,0}}      \expandafter\def\csname LT2\endcsname{\color[rgb]{0,0,1}}      \expandafter\def\csname LT3\endcsname{\color[rgb]{1,0,1}}      \expandafter\def\csname LT4\endcsname{\color[rgb]{0,1,1}}      \expandafter\def\csname LT5\endcsname{\color[rgb]{1,1,0}}      \expandafter\def\csname LT6\endcsname{\color[rgb]{0,0,0}}      \expandafter\def\csname LT7\endcsname{\color[rgb]{1,0.3,0}}      \expandafter\def\csname LT8\endcsname{\color[rgb]{0.5,0.5,0.5}}    \else
            \def\colorrgb#1{\color{black}}      \def\colorgray#1{\color[gray]{#1}}      \expandafter\def\csname LTw\endcsname{\color{white}}      \expandafter\def\csname LTb\endcsname{\color{black}}      \expandafter\def\csname LTa\endcsname{\color{black}}      \expandafter\def\csname LT0\endcsname{\color{black}}      \expandafter\def\csname LT1\endcsname{\color{black}}      \expandafter\def\csname LT2\endcsname{\color{black}}      \expandafter\def\csname LT3\endcsname{\color{black}}      \expandafter\def\csname LT4\endcsname{\color{black}}      \expandafter\def\csname LT5\endcsname{\color{black}}      \expandafter\def\csname LT6\endcsname{\color{black}}      \expandafter\def\csname LT7\endcsname{\color{black}}      \expandafter\def\csname LT8\endcsname{\color{black}}    \fi
  \fi
    \setlength{\unitlength}{0.0500bp}    \ifx\gptboxheight\undefined      \newlength{\gptboxheight}      \newlength{\gptboxwidth}      \newsavebox{\gptboxtext}    \fi    \setlength{\fboxrule}{0.5pt}    \setlength{\fboxsep}{1pt}\begin{picture}(3440.00,2540.00)    \gplgaddtomacro\gplbacktext{      \csname LTb\endcsname      \put(747,595){\makebox(0,0)[r]{\strut{}82\%}}      \csname LTb\endcsname      \put(747,1030){\makebox(0,0)[r]{\strut{}84\%}}      \csname LTb\endcsname      \put(747,1465){\makebox(0,0)[r]{\strut{}86\%}}      \csname LTb\endcsname      \put(747,1900){\makebox(0,0)[r]{\strut{}88\%}}      \csname LTb\endcsname      \put(747,2335){\makebox(0,0)[r]{\strut{}90\%}}      \csname LTb\endcsname      \put(983,409){\makebox(0,0){\strut{}$1$}}      \csname LTb\endcsname      \put(1655,409){\makebox(0,0){\strut{}$2$}}      \csname LTb\endcsname      \put(2327,409){\makebox(0,0){\strut{}$4$}}      \csname LTb\endcsname      \put(2999,409){\makebox(0,0){\strut{}$8$}}    }    \gplgaddtomacro\gplfronttext{      \csname LTb\endcsname      \put(144,1465){\rotatebox{-270}{\makebox(0,0){\strut{}PCKh total}}}      \csname LTb\endcsname      \put(1991,130){\makebox(0,0){\strut{}Stacks}}      \csname LTb\endcsname      \put(2345,1134){\makebox(0,0)[r]{\strut{}\footnotesize HM}}      \csname LTb\endcsname      \put(2345,948){\makebox(0,0)[r]{\strut{}\footnotesize DSNT}}      \csname LTb\endcsname      \put(2345,762){\makebox(0,0)[r]{\strut{}\footnotesize DSNTr}}    }    \gplbacktext
    \put(0,0){\includegraphics{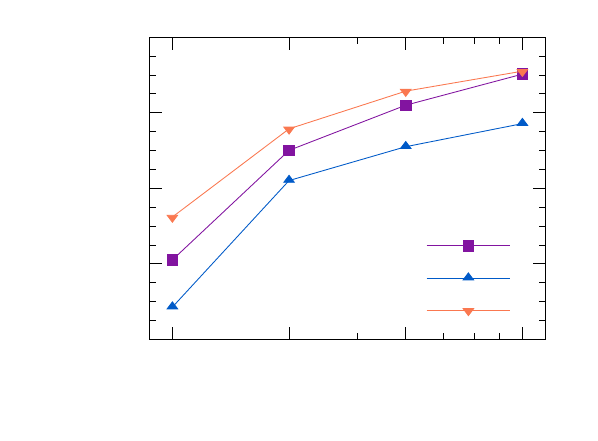}}    \gplfronttext
  \end{picture}\endgroup
\clearpage{}\vspace{-0.8cm}
\par\end{center}
\caption{\label{fig:hourglass}Varying output strategy and stack count for
hourglass \citep{newell2016stacked} models.}
\end{minipage}
\end{figure*}

\subsection{Model base}

We conducted experiments using two different fully convolutional model
architectures for the CNN portion of the coordinate regression network
(see \figref{arch_comparison}).
\begin{description}
\item [{ResNet}] The ResNet architecture \citep{he2016resnet} is well-known
for performing extremely well in classification tasks. We converted
ImageNet-pretrained ResNet models into fully convolutional networks
(FCNs) by removing the final fully connected classification layer.
Such models produce $7\times7$ px spatial heatmap outputs. However,
we were able to adjust the heatmap resolution of the FCN using dilated
convolutions, as proposed by \citet{yu2016dilated}. More specifically,
we change the convolution stride from 2 to 1 in one or more downsampling
stages, then use dilated convolutions in subsequent layers to preserve
the receptive field size. For each downsampling stage modified in
this way, the heatmap resolution increases by a factor of two.
\item [{Stacked hourglass}] The stacked hourglass architecture \citep{newell2016stacked}
is currently state-of-the-art for human pose estimation \citep{yang2017learning,chen2017advposenet,chou2017self,chu2017multi}.
The heatmap resolution of this architecture is $64\times64$ px.
\end{description}

\subsection{Output strategy}
\begin{description}
\item [{Heatmap matching (HM)}] We follow the specific technique used
by \citet{newell2016stacked}. MSE pixel-wise loss is applied directly
to the output of the FCN. During inference, numeric coordinates are
calculated based on the brightest pixel of the heatmap, with small
adjustments to the location made based on the brightness of adjacent
pixels.
\item [{Fully connected (FC)}] A softmax heatmap activation is applied
to the output of the FCN, followed by a fully connected layer which
produces numerical coordinates. The model is trained with Euclidean
loss.
\item [{DSNT}] Same as fully connected, but with our DSNT layer instead
of the fully connected layer.
\item [{DSNT with regularization (DSNTr)}] Same as DSNT, but with the
inclusion of a regularization term in the loss function. The method
of regularization we selected was Jensen-Shannon divergence with $\sigma_{t}=1$
and $\lambda=1$, which empirically performed best.
\end{description}

\subsection{Dataset and training}

We use the MPII human pose dataset \citep{andriluka20142d} to evaluate
the effectiveness of our DSNT layer on an important real-world task.
The dataset contains images of 28,883 people with up to 16 joint annotations
each, along with approximate person location and scale labels to facilitate
the cropping of single-person poses.

Samples from the dataset were augmented during training time using
the same scheme as \citet{newell2016stacked}, which consists of horizontal
flips, 75\%-125\% scaling, $\pm30$ degree rotation, and 60\%-140\%
channel-wise pixel value scaling. Since the test set labels are not
public, we evaluate on the fixed validation set used in \citep{tompson2015efficient}
and \citep{newell2016stacked}.

The models were optimized with RMSProp \citep{tieleman2012rmsprop}
using an initial learning rate of $2.5\times10^{-4}$. Each model
was trained for 120 epochs, with the learning rate reduced by a factor
of 10 at epochs 60 and 90 (an epoch is one complete pass over the
training set). Training was completed on single Maxwell-architecture
NVIDIA Titan X GPUs.

Our ResNet-based networks were trained using mini-batches of 32 samples
each, with the exception of highly memory-intensive configurations
(\eg ResNet-101@28px). The stacked hourglass models were trained
using mini-batches of 6 samples each. Our implementation code for
DSNT, written in PyTorch, is available online\footnote{https://github.com/anibali/dsntnn}.

\subsection{Results}

\begin{table*}
\begin{centering}
\caption{\label{tbl:pose_results}MPII human pose test set PCKh accuracies
and inference-time efficiency results.}
\bgroup\tabcolsep=0.1cm\centerline{ \begin{tabular}{l>{\raggedright}p{0.8cm}>{\raggedright}p{0.8cm}>{\raggedright}p{0.8cm}>{\raggedright}p{0.8cm}>{\raggedright}p{0.8cm}>{\raggedright}p{0.8cm}>{\raggedright}p{0.8cm}lll}
\toprule 
Method & Head & Shoul. & Elbow & Wrist & Hip & Knee & Ankle & Total & Time (ms){*} & Memory{*}\tabularnewline
\midrule
{\small{}\citet{tompson2015efficient}} & {\small{}96.1} & {\small{}91.9} & {\small{}83.9} & {\small{}77.8} & {\small{}80.9} & {\small{}72.3} & {\small{}64.8} & {\small{}82.0} & {\small{}-} & {\small{}-}\tabularnewline
{\small{}\citet{rafi2016efficient}} & {\small{}97.2} & {\small{}93.9} & {\small{}86.4} & {\small{}81.3} & {\small{}86.8} & {\small{}80.6} & {\small{}73.4} & {\small{}86.3} & {\small{}27.6$\pm$0.1} & {\small{}2768 MiB}\tabularnewline
{\small{}\citet{wei2016convolutional}} & {\small{}97.8} & {\small{}95.0} & {\small{}88.7} & {\small{}84.0} & {\small{}88.4} & {\small{}82.8} & {\small{}79.4} & {\small{}88.5} & {\small{}106.8$\pm$0.2} & {\small{}5832 MiB}\tabularnewline
{\small{}Bulat et al. \citep{bulat2016human}} & {\small{}97.9} & {\small{}95.1} & {\small{}89.9} & {\small{}85.3} & {\small{}89.4} & {\small{}85.7} & {\small{}81.7} & {\small{}89.7} & {\small{}41.3$\pm$0.2} & {\small{}1432 MiB}\tabularnewline
{\small{}\citet{newell2016stacked}} & {\small{}98.2} & {\small{}96.3} & {\small{}91.2} & {\small{}87.1} & {\small{}90.1} & {\small{}87.4} & {\small{}83.6} & {\small{}90.9} & {\small{}60.5$\pm$0.1} & {\small{}1229 MiB}\tabularnewline
{\small{}\citet{yang2017learning}} & \textbf{\small{}98.5} & \textbf{\small{}96.7} & \textbf{\small{}92.5} & \textbf{\small{}88.7} & \textbf{\small{}91.1} & \textbf{\small{}88.6} & \textbf{\small{}86.0} & \textbf{\small{}92.0} & {\small{}194.6$\pm$76.8} & {\small{}1476 MiB}\tabularnewline
\midrule 
{\small{}DSNTr ResNet-50@28px} & {\small{}97.8} & {\small{}96.0} & {\small{}90.0} & {\small{}84.3} & {\small{}89.8} & {\small{}85.2} & {\small{}79.7} & {\small{}89.5} & \textbf{\small{}18.6$\pm$0.5} & \textbf{\small{}636 MiB}\tabularnewline
\bottomrule
\end{tabular}}\egroup
\par\end{centering}
\centering{}{\small{}\smallskip{}
{*} Any test time data augmentations (horizontal flips, multi-scale)
were disabled for time and memory measurements.}{\small \par}
\end{table*}

The PCKh performance metric is the percentage of joints with predicted
locations that are no further than half of the head segment length
from the ground truth. As per the evaluation code provided by MPII,
we exclude the pelvis and thorax joints from the average total PCKh.

In order to compare the different approaches to coordinate regression,
we conducted a series of experiments with a ResNet-34-based network
(\figref{outstrat}). The heatmap matching achieved a very low PCKh
of 44\% at $7\times7$ px heatmap resolution, which falls outside
the bounds of the figure. As the resolution increases, the performance
of heatmap matching improves relative to the other approaches, which
is evidence of the quantization effects inherent to calculating coordinates
via a pixel-wise argmax. This demonstrates that heatmap matching is
not suitable for models which generate low-resolution heatmaps, whereas
DSNT is largely robust to heatmap size. At higher resolutions, the
fully connected approach performs worst. Our DSNT approach exhibits
good performance across all resolutions—even $7\times7$ px—due to
the predictions produced by DSNT not having precision dependent on
pixel size.

Regularization improves DSNT accuracy in all cases except the lowest
resolution, where boundary effects come into play (\ie a 1 pixel
standard deviation Gaussian drawn in a $7\times7$ px image is likely
to clip heavily, which adversely affects the DSNT calculation). Fully
connected output was found to be worse than heatmap matching at higher
resolutions, and worse than DSNT in general.

We conducted further experiments with ResNet-based \citep{he2016resnet}
models to evaluate the impact that depth has on performance. The results
in \figref{depth} suggest that higher heatmap resolution is beneficial
at any depth. However, the trade-off is that increasing resolution
with dilations has a large impact on memory consumption and computational
cost. For this reason, we could not train ResNet-101@56px. PCKh was
found to increase significantly with depth up until ResNet-50, with
only a slight gain observed when increasing the depth even further
to ResNet-101.

\begin{figure}
\begin{centering}
\clearpage{}\begingroup
  \makeatletter
  \providecommand\color[2][]{    \GenericError{(gnuplot) \space\space\space\@spaces}{      Package color not loaded in conjunction with
      terminal option `colourtext'    }{See the gnuplot documentation for explanation.    }{Either use 'blacktext' in gnuplot or load the package
      color.sty in LaTeX.}    \renewcommand\color[2][]{}  }  \providecommand\includegraphics[2][]{    \GenericError{(gnuplot) \space\space\space\@spaces}{      Package graphicx or graphics not loaded    }{See the gnuplot documentation for explanation.    }{The gnuplot epslatex terminal needs graphicx.sty or graphics.sty.}    \renewcommand\includegraphics[2][]{}  }  \providecommand\rotatebox[2]{#2}  \@ifundefined{ifGPcolor}{    \newif\ifGPcolor
    \GPcolortrue
  }{}  \@ifundefined{ifGPblacktext}{    \newif\ifGPblacktext
    \GPblacktexttrue
  }{}    \let\gplgaddtomacro\g@addto@macro
    \gdef\gplbacktext{}  \gdef\gplfronttext{}  \makeatother
  \ifGPblacktext
        \def\colorrgb#1{}    \def\colorgray#1{}  \else
        \ifGPcolor
      \def\colorrgb#1{\color[rgb]{#1}}      \def\colorgray#1{\color[gray]{#1}}      \expandafter\def\csname LTw\endcsname{\color{white}}      \expandafter\def\csname LTb\endcsname{\color{black}}      \expandafter\def\csname LTa\endcsname{\color{black}}      \expandafter\def\csname LT0\endcsname{\color[rgb]{1,0,0}}      \expandafter\def\csname LT1\endcsname{\color[rgb]{0,1,0}}      \expandafter\def\csname LT2\endcsname{\color[rgb]{0,0,1}}      \expandafter\def\csname LT3\endcsname{\color[rgb]{1,0,1}}      \expandafter\def\csname LT4\endcsname{\color[rgb]{0,1,1}}      \expandafter\def\csname LT5\endcsname{\color[rgb]{1,1,0}}      \expandafter\def\csname LT6\endcsname{\color[rgb]{0,0,0}}      \expandafter\def\csname LT7\endcsname{\color[rgb]{1,0.3,0}}      \expandafter\def\csname LT8\endcsname{\color[rgb]{0.5,0.5,0.5}}    \else
            \def\colorrgb#1{\color{black}}      \def\colorgray#1{\color[gray]{#1}}      \expandafter\def\csname LTw\endcsname{\color{white}}      \expandafter\def\csname LTb\endcsname{\color{black}}      \expandafter\def\csname LTa\endcsname{\color{black}}      \expandafter\def\csname LT0\endcsname{\color{black}}      \expandafter\def\csname LT1\endcsname{\color{black}}      \expandafter\def\csname LT2\endcsname{\color{black}}      \expandafter\def\csname LT3\endcsname{\color{black}}      \expandafter\def\csname LT4\endcsname{\color{black}}      \expandafter\def\csname LT5\endcsname{\color{black}}      \expandafter\def\csname LT6\endcsname{\color{black}}      \expandafter\def\csname LT7\endcsname{\color{black}}      \expandafter\def\csname LT8\endcsname{\color{black}}    \fi
  \fi
    \setlength{\unitlength}{0.0500bp}    \ifx\gptboxheight\undefined      \newlength{\gptboxheight}      \newlength{\gptboxwidth}      \newsavebox{\gptboxtext}    \fi    \setlength{\fboxrule}{0.5pt}    \setlength{\fboxsep}{1pt}\begin{picture}(4240.00,2540.00)    \gplgaddtomacro\gplbacktext{      \csname LTb\endcsname      \put(747,740){\makebox(0,0)[r]{\strut{}84\%}}      \csname LTb\endcsname      \put(747,1030){\makebox(0,0)[r]{\strut{}85\%}}      \csname LTb\endcsname      \put(747,1320){\makebox(0,0)[r]{\strut{}86\%}}      \csname LTb\endcsname      \put(747,1610){\makebox(0,0)[r]{\strut{}87\%}}      \csname LTb\endcsname      \put(747,1900){\makebox(0,0)[r]{\strut{}88\%}}      \csname LTb\endcsname      \put(747,2190){\makebox(0,0)[r]{\strut{}89\%}}      \csname LTb\endcsname      \put(849,409){\makebox(0,0){\strut{}$0$}}      \csname LTb\endcsname      \put(1235,409){\makebox(0,0){\strut{}$10$}}      \csname LTb\endcsname      \put(1620,409){\makebox(0,0){\strut{}$20$}}      \csname LTb\endcsname      \put(2006,409){\makebox(0,0){\strut{}$30$}}      \csname LTb\endcsname      \put(2391,409){\makebox(0,0){\strut{}$40$}}      \csname LTb\endcsname      \put(2777,409){\makebox(0,0){\strut{}$50$}}      \csname LTb\endcsname      \put(3162,409){\makebox(0,0){\strut{}$60$}}      \csname LTb\endcsname      \put(3548,409){\makebox(0,0){\strut{}$70$}}      \csname LTb\endcsname      \put(3933,409){\makebox(0,0){\strut{}$80$}}    }    \gplgaddtomacro\gplfronttext{      \csname LTb\endcsname      \put(144,1465){\rotatebox{-270}{\makebox(0,0){\strut{}PCKh total}}}      \csname LTb\endcsname      \put(2391,130){\makebox(0,0){\strut{}Inference time (ms)}}      \csname LTb\endcsname      \put(1531,747){\makebox(0,0){\strut{}\footnotesize \textcolor[HTML]{360054}{HG1}}}      \csname LTb\endcsname      \put(1750,1496){\makebox(0,0){\strut{}\footnotesize \textcolor[HTML]{360054}{HG2}}}      \csname LTb\endcsname      \put(2287,1843){\makebox(0,0){\strut{}\footnotesize \textcolor[HTML]{360054}{HG4}}}      \csname LTb\endcsname      \put(3521,2081){\makebox(0,0){\strut{}\footnotesize \textcolor[HTML]{360054}{HG8}}}      \csname LTb\endcsname      \put(3145,948){\makebox(0,0)[r]{\strut{}\footnotesize Hourglass, HM \citep{newell2016stacked}}}      \csname LTb\endcsname      \put(1165,1354){\makebox(0,0){\strut{}\footnotesize \textcolor[HTML]{00680e}{14px}}}      \csname LTb\endcsname      \put(1630,2027){\makebox(0,0){\strut{}\footnotesize \textcolor[HTML]{00680e}{28px}}}      \csname LTb\endcsname      \put(2828,1926){\makebox(0,0){\strut{}\footnotesize \textcolor[HTML]{00680e}{56px}}}      \csname LTb\endcsname      \put(3145,762){\makebox(0,0)[r]{\strut{}\footnotesize ResNet-50, DSNTr}}    }    \gplbacktext
    \put(0,0){\includegraphics{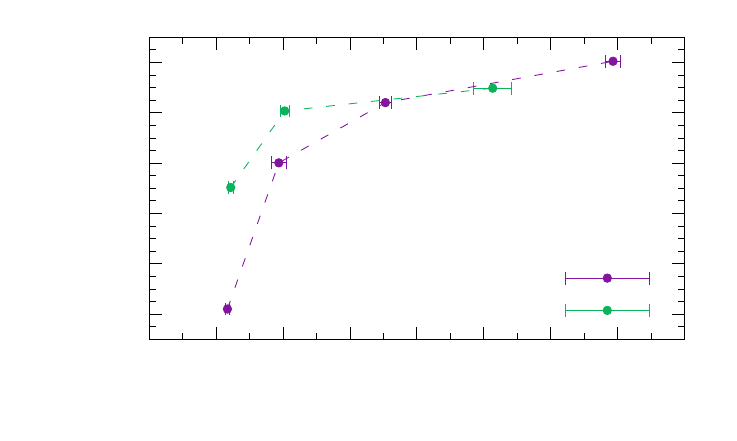}}    \gplfronttext
  \end{picture}\endgroup
\clearpage{}\vspace{-0.5cm}
\par\end{centering}
\caption{\label{fig:time}Validation accuracy vs inference time, closer to
the top-left is better. Labels show heatmap resolution (ResNet models)
or stack count (hourglass models).}
\end{figure}

In addition to ResNet, we also trained stacked hourglass networks
\citep{newell2016stacked}. Even though the stacked hourglass architecture
was developed using heatmap matching, we found that models trained
using DSNT with regularization achieved consistently better results
(\figref{hourglass}). Analysis of misclassified examples revealed
that DSNT was less accurate for predicting edge case joints that lie
very close to the image boundary, which is expected due to how the
layer works.

\figref{time} directly compares stacked hourglass networks trained
with heatmap matching and our ResNet-based networks trained with DSNT
and regularization. Although the 8-stack hourglass network was found
to have the highest overall accuracy, the ResNet-based models were
found to be much faster with only modest concessions in terms of accuracy.
For instance, ResNet-50@28px has 8\% fewer parameters, requires less
than half of the memory during training, and is over $3\times$ faster
at inference than HG8, whilst still achieving \textasciitilde{}99\%
of the PCKh score.

\begin{figure}
\begin{centering}
\clearpage{}\begingroup
  \makeatletter
  \providecommand\color[2][]{    \GenericError{(gnuplot) \space\space\space\@spaces}{      Package color not loaded in conjunction with
      terminal option `colourtext'    }{See the gnuplot documentation for explanation.    }{Either use 'blacktext' in gnuplot or load the package
      color.sty in LaTeX.}    \renewcommand\color[2][]{}  }  \providecommand\includegraphics[2][]{    \GenericError{(gnuplot) \space\space\space\@spaces}{      Package graphicx or graphics not loaded    }{See the gnuplot documentation for explanation.    }{The gnuplot epslatex terminal needs graphicx.sty or graphics.sty.}    \renewcommand\includegraphics[2][]{}  }  \providecommand\rotatebox[2]{#2}  \@ifundefined{ifGPcolor}{    \newif\ifGPcolor
    \GPcolortrue
  }{}  \@ifundefined{ifGPblacktext}{    \newif\ifGPblacktext
    \GPblacktexttrue
  }{}    \let\gplgaddtomacro\g@addto@macro
    \gdef\gplbacktext{}  \gdef\gplfronttext{}  \makeatother
  \ifGPblacktext
        \def\colorrgb#1{}    \def\colorgray#1{}  \else
        \ifGPcolor
      \def\colorrgb#1{\color[rgb]{#1}}      \def\colorgray#1{\color[gray]{#1}}      \expandafter\def\csname LTw\endcsname{\color{white}}      \expandafter\def\csname LTb\endcsname{\color{black}}      \expandafter\def\csname LTa\endcsname{\color{black}}      \expandafter\def\csname LT0\endcsname{\color[rgb]{1,0,0}}      \expandafter\def\csname LT1\endcsname{\color[rgb]{0,1,0}}      \expandafter\def\csname LT2\endcsname{\color[rgb]{0,0,1}}      \expandafter\def\csname LT3\endcsname{\color[rgb]{1,0,1}}      \expandafter\def\csname LT4\endcsname{\color[rgb]{0,1,1}}      \expandafter\def\csname LT5\endcsname{\color[rgb]{1,1,0}}      \expandafter\def\csname LT6\endcsname{\color[rgb]{0,0,0}}      \expandafter\def\csname LT7\endcsname{\color[rgb]{1,0.3,0}}      \expandafter\def\csname LT8\endcsname{\color[rgb]{0.5,0.5,0.5}}    \else
            \def\colorrgb#1{\color{black}}      \def\colorgray#1{\color[gray]{#1}}      \expandafter\def\csname LTw\endcsname{\color{white}}      \expandafter\def\csname LTb\endcsname{\color{black}}      \expandafter\def\csname LTa\endcsname{\color{black}}      \expandafter\def\csname LT0\endcsname{\color{black}}      \expandafter\def\csname LT1\endcsname{\color{black}}      \expandafter\def\csname LT2\endcsname{\color{black}}      \expandafter\def\csname LT3\endcsname{\color{black}}      \expandafter\def\csname LT4\endcsname{\color{black}}      \expandafter\def\csname LT5\endcsname{\color{black}}      \expandafter\def\csname LT6\endcsname{\color{black}}      \expandafter\def\csname LT7\endcsname{\color{black}}      \expandafter\def\csname LT8\endcsname{\color{black}}    \fi
  \fi
    \setlength{\unitlength}{0.0500bp}    \ifx\gptboxheight\undefined      \newlength{\gptboxheight}      \newlength{\gptboxwidth}      \newsavebox{\gptboxtext}    \fi    \setlength{\fboxrule}{0.5pt}    \setlength{\fboxsep}{1pt}\begin{picture}(4240.00,2540.00)    \gplgaddtomacro\gplbacktext{      \csname LTb\endcsname      \put(747,595){\makebox(0,0)[r]{\strut{}20\%}}      \csname LTb\endcsname      \put(747,885){\makebox(0,0)[r]{\strut{}30\%}}      \csname LTb\endcsname      \put(747,1175){\makebox(0,0)[r]{\strut{}40\%}}      \csname LTb\endcsname      \put(747,1465){\makebox(0,0)[r]{\strut{}50\%}}      \csname LTb\endcsname      \put(747,1755){\makebox(0,0)[r]{\strut{}60\%}}      \csname LTb\endcsname      \put(747,2045){\makebox(0,0)[r]{\strut{}70\%}}      \csname LTb\endcsname      \put(747,2335){\makebox(0,0)[r]{\strut{}80\%}}      \csname LTb\endcsname      \put(1030,409){\makebox(0,0){\strut{}$1024$}}      \csname LTb\endcsname      \put(1937,409){\makebox(0,0){\strut{}$2048$}}      \csname LTb\endcsname      \put(2845,409){\makebox(0,0){\strut{}$4096$}}      \csname LTb\endcsname      \put(3752,409){\makebox(0,0){\strut{}$8192$}}    }    \gplgaddtomacro\gplfronttext{      \csname LTb\endcsname      \put(144,1465){\rotatebox{-270}{\makebox(0,0){\strut{}PCKh total}}}      \csname LTb\endcsname      \put(2391,130){\makebox(0,0){\strut{}Number of samples}}      \csname LTb\endcsname      \put(3145,1134){\makebox(0,0)[r]{\strut{}\footnotesize HM}}      \csname LTb\endcsname      \put(3145,948){\makebox(0,0)[r]{\strut{}\footnotesize FC}}      \csname LTb\endcsname      \put(3145,762){\makebox(0,0)[r]{\strut{}\footnotesize DSNTr}}    }    \gplbacktext
    \put(0,0){\includegraphics{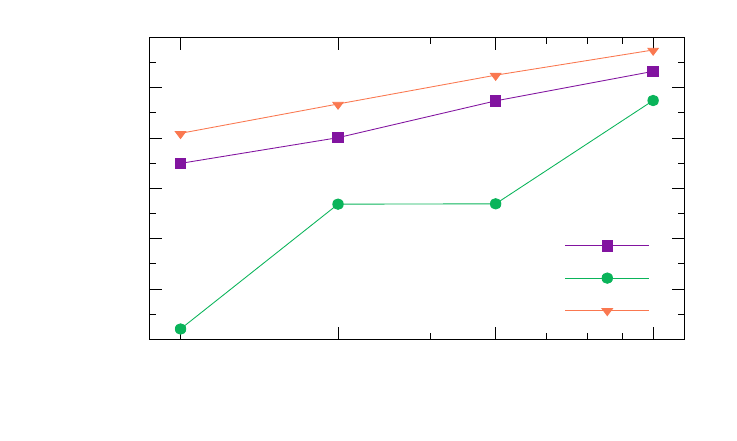}}    \gplfronttext
  \end{picture}\endgroup
\clearpage{}\vspace{-0.5cm}
\par\end{centering}
\caption{\label{fig:nsamples}Varying number of training samples (no augmentation)
for ResNet-34@28px models.}
\end{figure}

Spatial generalization was tested by training models with a restricted
training set size and no data augmentation. \figref{nsamples} shows
that fully connected output exhibits very poor spatial generalization,
achieving the extremely low PCKh score of 22\% when trained on 1024
samples. On the other hand, both DSNT and heatmap matching perform
much better with fewer samples, indicating better generalization.

Finally, we evaluated our ResNet-50@28px DSNTr model on the test set.
The results in \tblref{pose_results} show that our solution, using
a much smaller and simpler model (ResNet-50), was able to achieve
accuracy competitive with more complex models. A consequence of using
a smaller model is that ResNet-50@28px infers significantly faster
and uses less memory than all other methods shown in the table. Note
that we determined the running time and memory usage of the other
methods by downloading pretrained models.

\section{Conclusion}

There are multiple possible approaches to using CNNs for numerical
coordinate regression tasks, each of which affects the behavior of
the model in different ways—a fully connected output layer reduces
spatial generalization, and heatmap matching introduces issues with
differentiability and quantization. In contrast, our proposed DSNT
layer can be used to adapt fully convolutional networks for coordinate
regression without introducing these problems. We have shown that
models built with DSNT can achieve competitive results on real human
pose data without complex task-specific architectures, forming a strong
baseline. Such models also offer a better accuracy to inference speed
trade-off when compared with stacked hourglass models.

Interesting directions for future work are to integrate DSNT with
complex pose estimation approaches (\eg adversarial training \citep{chou2017self,chen2017advposenet}),
or to use DSNT as an internal layer for models where intermediate
coordinate prediction is required (\eg Spatial Transformer Networks
\citep{jaderberg2015spatial}).

\bibliographystyle{IEEEtranN}
\phantomsection\addcontentsline{toc}{section}{\refname}\bibliography{references}

\end{document}